# ROBUST FUZZY CORNER DETECTOR

*Erik Cuevas, *Daniel Zaldivar, *Marco Pérez-Cisneros +Edgar Sánchez and ⁻Marte Ramírez-Ortegón

*Departamento de Ciencias Computacionales*
*Universidad de Guadalajara, CUCEI*
*Av. Revolución 1500, Guadalajara, Jal, México*
*{erik.cuevas, daniel.zaldivar, marco.perez}@cucei.udg.mx*
*Telephone: +52 33 1378 5900, Ext: 7715*
+*CINVESTAV-Unidad Guadalajara*
*Av. Científica 1145, Colonia el Bajío, Zapopan, 45010, Jal, México.*
⁻*Freie Universität Berlin*
*Takustrs 9, PLZ 14195, Berlin, Germany*

ABSTRACT— Reliable corner detection is an important task in determining the shape of different regions within an image. Real-life image data are always imprecise due to inherent uncertainties that may arise from the imaging process such as defocusing, illumination changes, noise, etc. Therefore, the localization and detection of corners has become a difficult task to accomplish under such imperfect situations. On the other hand, Fuzzy systems are well known for their efficient handling of impreciseness and incompleteness, which make them inherently suitable for modelling corner properties by means of a rule-based fuzzy system. The paper presents a corner detection algorithm which employs such fuzzy reasoning. The robustness of the proposed algorithm is compared to well-known conventional corner detectors and its performance is also tested over a number of benchmark images to illustrate the efficiency of the algorithm under uncertainty.

## 1. INTRODUCTION

The human visual system has a highly developed capability for detecting many classes of patterns including visually significant arrangements of image elements. From the psychovisual aspect, points representing high curvature are one of the dominant classes of patterns that play an important role in almost all real life image analysis applications [1, 2, 3]. These points encode a significant amount of shape information. Corners are generally formed at the junction of different edge segments which may be the meeting (or crossing) of two edges. Cornerness of an edge segment depends solely on the curvature formed at the meeting point of two line segments. Corner detection is one of the fundamental tasks in computer vision and it can be regarded as a special type of feature segmentation. Extracted corners can be used for measurement and/or recognition purposes. A large number of algorithms already exist in the literature. In particular, corner detection on gray level images can be classified into two main approaches. In the first approach, the gray level image is first converted into its binary version for extraction of boundaries using some thresholding technique. After a successful extraction of boundaries, the corners or the high curvature points are detected using directional codes or other polygonal approximation techniques [4]. In the second approach, the gray level image is taken directly as an input for corner detection. In this paper, the discussion is restricted to the second approach only. Most of the general-purpose detectors based on gray level, use either a topology-based or an auto-correlation- based approach. Topology based corner detectors, mainly use gradients and surface curvature to define the measure of cornerness. Points are marked as corners, if the value of cornerness exceeds some predefined threshold condition. Alternatively a measure of curvature can be obtained using auto-correlation [5–9].

There exits several classical corner detection algorithms for estimating corner points. Such detectors are based on a local structure matrix which consists on the first partial derivatives of the intensity function. An clear example is the Harris feature point detector [10], which is based on a comparison: the measure of the corner strength - which is defined by the method and is based on a local structure matrix - is compared to an appropriately chosen concrete threshold. Another well known corner detector is the SUSAN (Smallest Univalue Segment Assimilating Nucleus) detector which is based on brightness comparison [11]. It does not depend on image derivatives. The SUSAN area will reach a minimum while the nucleus lies on a corner point. The effectiveness of the above mentioned algorithms is acceptable. Recent studies such as [12] demonstrate that the Harris corner detector performs better for several circumstances in comparison to the SUSAN algorithm.

Data from natural images are always imprecise and noisy due to inherent uncertainties that may arise from the imaging process (such as defocusing, wide variations of illuminations, etc.). As a result, precise localization and detection of corners become difficult under such imperfect situations. On the other hand, Fuzzy systems are well known for efficiently handling of impreciseness and incompleteness [13, 14, 15] due to imperfection of data. Therefore it may result reasonable to model corner properties using a fuzzy rule-based system as they have been successfully applied to image processing in a wide variety of applications [16-18]. This paper seeks to contribute to enhance the application of fuzzy logic to image processing, just as it has been proposed in [19]. The method adopts a template-based rule-driven procedure and has been specifically developed to deal with topics related to image processing purposes. This method is able to address many different processing tasks [20-22] and to produce better results than classical methods when applied to some critical issues such as noise [20,23,24]. Only few fuzzy approaches have specifically addressed the problem of corner detection for general purposes. Banerjee & Kundu have proposed in [25] an algorithm to extract significant gray level corner points. The measure of cornerness in each point is computed by means of the fuzzy edge strength and the gradient direction. Different corner fuzzy-sets are obtained by considering different threshold values from the fuzzy edge map. However, the algorithm's main drawback is that it uses several feature detectors which operate at different stages, yielding a high computational load. On the other hand, Várkonyi-Kóczy have proposed in [26], a fuzzy corner detector that employs a local structure matrix. It builds a continuous transient between the localized and not localized corner points. The algorithm uses a fuzzy pre-filter that improve the quality of the image under process. Despite both fuzzy approaches show a good performance, they demand an expensive computing load in comparison to other classical algorithms such as the Harris method or SUSAN.

This paper presents a new robust algorithm to extract significant gray level corner points. The method is derived from a fuzzy-rule approach which aims to detect corners even under complex conditions. In the proposed approach, the measure of "cornerness" for each pixel in the image is computed by fuzzy rules (represented as templates) which are applied to a set of pixels belonging to a rectangular window. As the algorithm scans each pixel of the image at a time, a new pixel of the resulting image is generated by fuzzy reasoning. Hence, the possible uncertainty contained in the window-neighborhood is handled by using an appropriate rule base (template set). Experimental evidence shows the effectiveness of such method for detecting corners under several conditions. A comparison between one state-of-the-art Fuzzy-based method [25] and the Harris algorithm [10] demonstrates the performance of the proposed method.

The paper is organized as follows: Section 2 briefly describes the mathematical approach and the fuzzy model used in this work. Section 3 describes the features extraction process while Section 4 describes the fuzzy corner extraction process. On the other hand, Section 5 describes the experimental results while Section 6 offers some conclusions about the development and performance of this technique.

## 2. FUZZY RULE-BASED SYSTEM

### *2.1 Fuzzy System*

Most of the approaches for corner detection are easy to implement and demand a low computational load. However, their effective operation largely relies on the fact that noisiness must be limited. In this section, a more robust technique is proposed. The new procedure is set to deliver a better performance for noisy environments. The fuzzy system is simple to implement and still fast in computation if it is compared to some existing fuzzy methods [25,26]. Also, it can be easily extended to detect other features. In the proposed approach, the fuzzy rules are applied to a set of pixels belonging to a rectangular $N \times N$ window (usually 3x3 pixels), where the gray-level differences between the center pixel and its surrounding pixels are computed and stored within matrix $E$ as follows:

$$E = \begin{bmatrix} p_{m,n} - p_{m-1,n-1} & p_{m,n} - p_{m-1,n} & p_{m,n} - p_{m-1,n+1} \\ p_{m,n} - p_{m,n-1} & 0 & p_{m,n} - p_{m,n+1} \\ p_{m,n} - p_{m+1,n-1} & p_{m,n} - p_{m+1,n} & p_{m,n} - p_{m+1,n+1} \end{bmatrix} \quad (1)$$

where $m$ and $n$ represent the coordinates of the central pixel. If the neighborhood is a homogenous region, then $E$ contains values near zero. In the case of corners, the matrix $E$ possesses a specific configuration depending on the corner type. These divide $E$ in two connected regions, one with positive (pixel type ***A***) and another with negative (pixel type ***B***) difference values (see Figure 1). The reasoning structure uses two different types of rules: the ***THEN-rules*** and the ***ELSE-rules*** (don't care conditions) respectively. Each THEN-rule includes a determined pixel configuration as antecedent and only one pixel as consequent. Antecedents are related to a corner existence test and the consequent to its presence or absence. The rule-base gathers many fuzzy rules (THEN-rules) and only one ELSE-rule (i.e. do-not-care rule). Therefore only relevant rules (i.e. configurations) are formulated as THEN-rules while other not important configurations may be handled as a group of ELSE-rules. The set of THEN-rules lies on the very core of the algorithm. The rules must deliver successful structure detection, i.e. corners in this case, while still cancelling other inconsistencies such as noise. Such tradeoff may be solved by using a reduced set of rules (configurations) which in turn represent the minimum number in order to coherently detect the structure as it is required by a given application. Such procedure allows dealing with noisy pixels or imprecision.

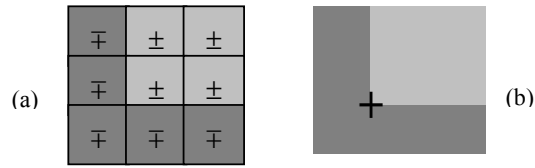

**Fig. 1.** Region shaping with respect to gray level differences: (a) the resulting template and (b) the real corner that originates the template.

The proposed corner detector considers twelve THEN-rules that represent the same number of possible corner configurations and only one ELSE-rule as it is graphically explained by Fig. 2. It may be also possible to consider some other corner configurations. However it may reduce the algorithm's ability to deal with noise or uncertainty [19,20,24].

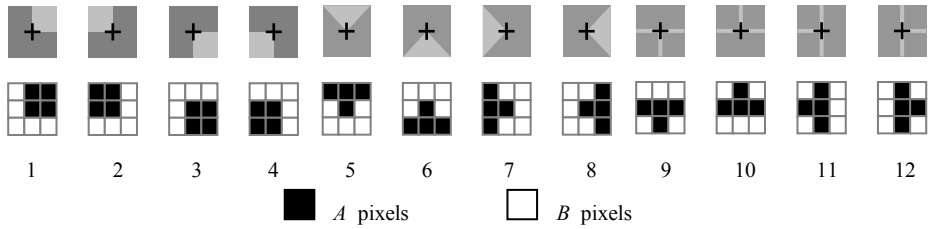

**Fig. 2.** Different corner cases to be considered for building the fuzzy rules. The image region containing the corner is shown in the upper section while the resulting 3x3 template is shown below each case.

Despite using a reduced rule base, the performance in the detection process can be considered acceptable when it is compared to other algorithms solving the same task. Each rule has the following form:

**If** the corner structure in $E$ possesses positive elements

**and** the opposite region possesses negative elements,

**then** the pixel represent a corner, (2)

**else** the pixel does not represent a corner

The principle can be explained as follows: If one region of the neighborhood, according to any of the twelve cases, contains positive/negative differences with respect to the center pixel, and if any other region contains the opposite (negative/positive) differences with respect to the center pixel, then the center pixel is a corner (see Fig. 2). The procedure can be considered as the evaluation of each one of the 12 different THEN-rules (configurations), yielding two auxiliary matrices $E^p$ and $E^n$ as follows:

$$E^p(i,j) = \begin{cases} 1 & \text{if } E(i,j) \geq 0 \\ 0 & \text{else} \end{cases} \qquad E^n(i,j) = \begin{cases} 1 & \text{if } E(i,j) < 0 \\ 0 & \text{else} \end{cases} \tag{3}$$

where $i, j$ represents de row and column of the matrix $E$ ($i, j \in (1,2,3)$), Eq. 1. For the case that all elements of $E^p / E^n$ are ones (meaning all elements of $E(i,j)$ are positives or negatives), it is possible to construct regions $A$ and $B$ within the window-neighbourhood according to the existing relative differences. Thus the values of $E^p$ and $E^n$ can be recalculated as follows:

$$E^p(i,j) = \begin{cases} 1 & \text{if } E(i,j) \leq t_h \\ 0 & \text{else} \end{cases} \qquad E^n(i,j) = \begin{cases} 1 & \text{if } E(i,j) > t_h \\ 0 & \text{else} \end{cases} \tag{4}$$

For all the elements of $E^p$ being ones, and

$$E^p(i,j) = \begin{cases} 1 & \text{if } E(i,j) \geq -t_h \\ 0 & \text{else} \end{cases} \qquad E^n(i,j) = \begin{cases} 1 & \text{if } E(i,j) < -t_h \\ 0 & \text{else} \end{cases} \tag{5}$$

For all the elements of $E^n$ being ones, $t_h$ is a threshold that controls the sensitivity of the considered differences. Typical values for $t_h$ normally fall into the interval (5,35). The lowest value of 5 would yield a higher detector's sensitivity which may detect a great number of corners corresponding to noisy intensity changes which are commonly found in images. On the other hand, a maximum value of 35 would detect corners matching to a significant difference between several objects in the structure, i.e. object whose pixels may be considered as being connected. Although the selection of the best value for $t_h$ clearly depends on the particular application, a good compromise can be obtained by taking a value on approximately half the overall interval, i.e. $t_h = 20$. The membership values $\mu_c(m,n)$ (where $c = 1,2,\ldots,12$) are computed depending on the corner types (see Fig. 2). According to [33], such values represent the antecedents of each employed THEN-rule. They can be calculated as follows:

$$\mu_c(m,n) = \frac{1}{20} \max \left[ \left( \sum_{ij \in A} E^p(i,j) \right) \cdot \left( \sum_{ij \in B} E^n(i,j) \right), \left( \sum_{ij \in B} E^p(i,j) \right) \cdot \left( \sum_{ij \in A} E^n(i,j) \right) \right] \tag{6}$$

Eq. (6) considers a normalization factor equal to 20 which represents the maximum possible value, i.e. the highest product of the multiplication among the pixels between $E^p$ and $E^n$. Hence, the membership value $\mu_c(m,n)$ falls between 0 and 1. Eq. (6) can be considered as the numerical implementation of the generic rule previously defined by Eq. 2. If Rule 1 (case 1) is considered as an example, the expressions corresponding to Eq. (6) would thus be:

$$\begin{aligned} \sum_{ij \in A} E^p(i,j) &= E^p(1,2) + E^p(1,3) + E^p(2,2) + E^p(2,3) \\ \sum_{ij \in B} E^n(i,j) &= E^n(1,1) + E^n(2,1) + E^n(3,1) + E^n(3,2) + E^n(3,3) \end{aligned} \tag{7}$$

$$\sum_{ij \in B} E^p(i,j) = E^p(1,1) + E^p(2,1) + E^p(3,1) + E^p(3,2) + E^p(3,3)$$

$$\sum_{ij \in A} E^n(i,j) = E^n(1,2) + E^n(1,3) + E^n(2,2) + E^n(2,3)$$

Analogously to Eq. (7), membership values $\mu_2(i,j),...,\mu_{12}(i,j)$ for other rules (cases) can be calculated. Finally, the 12 fuzzy rules can be added into a single fuzzy value using the ***max*** (maximum) operator. The final fuzzy value represents the linguistic meaning of cornerness yielding:

$$\mu_{cornerness}(m,n) = \max(\mu_1(m,n), \mu_2(m,n),...,\mu_{12}(m,n)) \tag{8}$$

The pixels whose value $\mu_{cornerness}(m,n)$ are near to one, belong to a feature similar to a corner, while values near to zero would represent any other feature.

## 2.2 Robustness

This kind of corner detection clearly differs from other classical procedures in several ways. Conventional corner detectors look usually for the explicit corner location by means of detecting the zero-crossing of derivatives in different directions. On the contrary the proposed approach detects the entire area where the corner could lie. In particular, gradient-based methods are normally highly sensitive to the noise in real images and being mainly affected by the impulsive noise. Also, most of the corner detection algorithms incorporate several pre-filters [27,10,11,1], which allow attenuation but do not eliminate impulsive noise. On the other hand, fuzzy detectors allow corner marking despite noisy environments either by implementing fuzzy pre-filtering that eliminates uncertainty on the image or by incorporating fuzzy sets for modeling imprecision [25]. The method presented in this paper considers vagueness due to noise and grayness ambiguity to be handled by the fuzzy rules introduced in Eq. (2). Considering the image in Figure 3, a pixel holding a different gray value from its neighbors is located within a homogeneous region. This situation can be considered as impulsive noise.

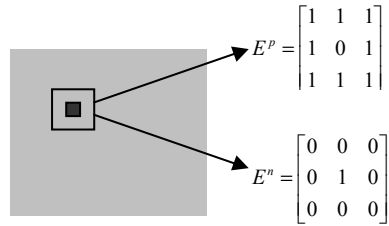

**Fig. 3.** The effect of the impulsive noise in matrices $E^+$ and $E^-$. Matrices $E^+$ and $E^-$ would contain only ones or zeros depending on the gray-level difference.

Under these circumstances, matrices $E^p$ and $E^p$ would contain only ones or zeros depending on the gray-level difference. Therefore, the values used to calculate the membership functions in Eq. (6), for any of the twelve cases, would yield

$$\left(\sum_{ij \in Rp} E^p(i,j)\right) \cdot \left(\sum_{ij \in Rn} E^p(i,j)\right) \approx 0 \qquad \left(\sum_{ij \in Rn} E^p(i,j)\right) \cdot \left(\sum_{ij \in Rp} E^n(i,j)\right) \approx 0 \tag{9}$$

Now, considering the values from Eq. (9) and a noisy pixel, the resulting value of its cornerness can be calculated by Eq. (8) as $\mu_{cornerness}(i,j) \approx 0$. The impulsive noise is thus classified by the fuzzy system as a homogeneous region. In the same way, the central pixel would not be marked as corner for cases not considered in Figure 1 which normally represent noisy configurations. It is mainly because the inference system works with ELSE-rules.

### *2.3 Corner selection*

In order to detect corners, it would be enough to choose an appropriate threshold $t_c$. If $\mu_{cornerness}(m,n) \geq t_c$, then the pixel $p_{m,n}$ can be assumed as such. Under these assumptions, the value $t_c$ must be selected as close to 1 as it is likely to assure that pixel $p_{m,n}$ may be a corner. However, a more convenient approach is to choose a small threshold value $t_c$ whose value allows detecting a wider number of corners despite a higher uncertainty. The corner selection process can therefore be explained as follows: For each pixel, if $\mu_{cornerness}(m,n) \geq t_c$, a neighborhood of $H \times H$ dimension is established around it (commonly $H > N$). The pixel $p_{m,n}$ is thus selected as a corner if its value $\mu_{cornerness}(m,n)$ is maximum within the neighborhood $H \times H$, otherwise it does not represent a corner point. Figure 4 shows a selection example, where $\mu_{cornerness}(m,n)$ represents the cornerness of the pixel currently under evaluation, by assuming $\mu_{cornerness}(m,n) \geq t_c$. Inside the window $H \times H$ that has been established around it, there exist other two pixels $p_{i,j}$ and $p_{i',j'}$, whose values $\mu_{cornerness}(i,j)$ and $\mu_{cornerness}(i',j')$ are lower than $\mu_{cornerness}(m,n)$. Therefore, a point $p_{m,n}$ can thus be considered as a corner within the image.

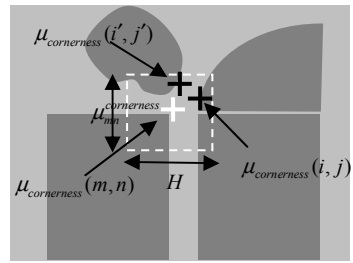

**Fig. 4.** Neighborhood method for corner selection.

### 3. EXPERIMENTAL RESULTS

Different sorts of images have been tested in order to analyze the performance of the method for corner detection. Such benchmark set includes image alterations such as blurring, illumination change, impulsive noise etc. Table 1 presents the parameters of the proposed algorithm used in this paper. Once they have been determined experimentally, they are kept for all the test images through all experiments.

| $t_h$ | $t_c$ | $H$ |
|---|---|---|
| 20 | 0.7 | 10 |

**Table 1.** Parameter setup for the proposed corner detector

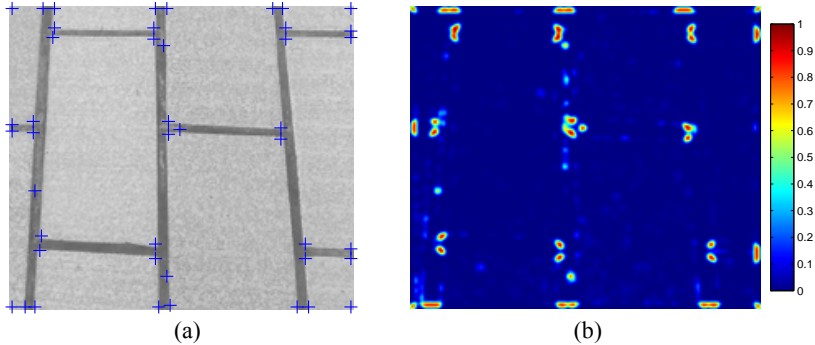

(a) (b)

**Fig. 5.** (a) Detected corners using the proposed approach, and (b) values of $\mu_{cornerness}(m,n)$

First, Figure 5b shows the value of $\mu_{cornerness}(m,n)$, as it is computed by the fuzzy system according to Eq. (9) to detect corners in a real image. In Figure 5a, the blue crosses represent the corners obtained using the corner selection procedure explained in sub-section 2.3. Figure 6 shows the algorithm´s performance on different image conditions such as the case with variable illumination and blurring. Figures 6(a)-(b) present the performance of the fuzzy corner detector as it is applied to over-exposed and over-illuminated images. The effect of high illumination on the images was made by applying a linear transformation of the form $I(i,j)+80$. On the other hand, Figures 6(c)-(d) show the effectiveness of the proposed detector using low-illuminated or sub-exposed images. Such effect was made by another linear transformation: $I(i,j)-40$. The images in Figures 6(e)-(f) illustrate the sensitivity of the fuzzy detector to blurring. Such steamed up effect was made by applying a low-pass filter to the original images, with a 5x5 kernel.

From results shown in Figure 6, it can be observed as the fuzzy detector exhibits immunity to changes in illumination, see for instance Figures 6(a)-(b) and 6(c)-(d). However, it also shows sensitivity to blurring in Figures 6(e)-(f). For the case of blurring images, the detector is able to find all the corners over the simulated image in 6(e). The latter figure exhibits low distortion in the homogeneous gray levels within the image as a consequence of the filter operation. On the other hand, some sensitivity may be lost while applying the detector to the real image shown in Figure 6(f). Moreover, after applying distortion to the image, several points that do not belong to a corner as such have been wrongly marked as corners. Despite all previous comments, the fuzzy detector was able to detect in Figure 6(f) the corners which delimit the object´s shape. This is not a common feature of other corner detectors [9-12].

## 4. PERFORMANCE COMPARISON

A variety of quantitative evaluation methods for corner detection algorithms have been proposed in the literature [12, 28, 29]. Following the criteria in [29], the performance analysis considers the Harris algorithm [10], the fuzzy method presented by Banerjee & Kundu [25] and the approach proposed in this paper. A quantitative comparison over three criteria is presented: stability, noise immunity and computational effort. The study aims analyze the performance objectively. The parameters for each detector algorithm are set as follows: For the Harris algorithm, the gradient operators [-2 -1 0 1 2] and [-2 -1 0 1 2]$^T$ are set in directions *u* and *v* separately. The Gaussian smoothing filter employs a Gaussian window function of size 7×7 and a standard deviation of 2 with *k*=0.06. The parametric setup appears as the best set following data in [12] and considering lots of hand tuning experiments. For the fuzzy method proposed by Banerjee & Kundu, the parameter are set following guidelines from [25], with a Gaussian window function of size 3×3 and a standard deviation of 2, $\mu_d(P) \geq 0.9$ and $T_h = 0.2$. Finally the parameters of the proposed approach are set according to the Table 1.

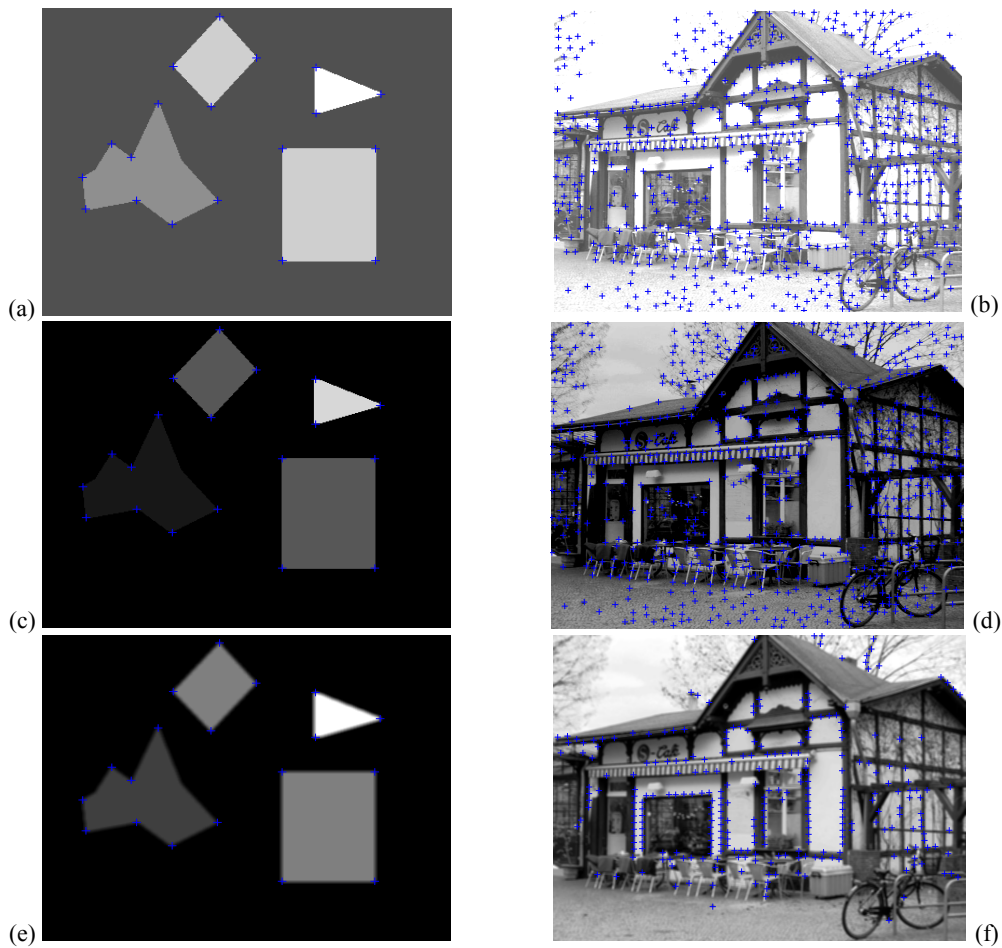

(a) (b) (c) (d) (e) (f)

**Fig. 6.** Performance of the fuzzy corner detector over different conditions on the image: (a)-(b) over-exposition or high illumination, (c)-(d) sub-exposition or low illumination and (e)-(f) blurring.

*4.1 Stability criterion*

Two frames in an image sequence are processed by the algorithm to detect corners. If the corner's positions are unchanged from one frame to the next one, the algorithm can be regarded as stable. However, the gray-level value of each pixel would normally vary in actual images because of several factors affecting the image. If the algorithm is applied to a given image, then it cannot be assured the number and position of all detected corners would be exactly the same. Therefore, absolute stability is almost non-existent. A factor $\eta$ to measure the stability of a corner algorithm can be defined as follows:

$$\eta = \frac{A_1 \cap A_2}{\min(|A_1|,|A_2|)} \times 100\%, \quad (10)$$

where $A_1$ and $A_2$ representing the corner sets for the first and the second frame respectively (the intersection operator $\cap$ stands for common corners); $|A_i|$ represents the number of elements in $A_i$ set and the overall numerator holds the number of corresponding corners in two frames. From Eq. (10), it can be concluded that a greater $\eta$ yields a more stable corner detection algorithm. Fifty pairs of images holding different contrast and brightness levels are gathered in order to compare the proposed fuzzy detector and other classic methods. Figure 7a shows the comparison with respect to the stability factor, where the horizontal axis represents the image pair number and the vertical axis represents the value of such stability factor. The average stability factor of Harris detector is 75%, while the fuzzy method Banerjee & Kundu holds 70% and the proposed fuzzy detector shows 83%.

*4.2 Noise immunity*

Noise immunity is measured by factor $\rho$ which it can be defined as follows:

$$\rho = \frac{B_1 \cap B_2}{\max(|B_1|,|B_2|)} \times 100\%, \quad (11)$$

where $B_1$ is the corner set of the original image and $B_2$ is the corner set of the image with added noise. In this case, the maximum operator seeks to consider that false corners have been added as a result of additive noise. As $\rho$ increases, it can be assumed that the algorithm's ability to avoid noisy corners is stronger. One experiment is focus on comparing such noise immunity among methods. Fifty images with 10% of added impulsive noise are considered. Figure 7b shows the noise immunity factor, with the Harris detector showing 9%, the fuzzy method Banerjee & Kundu holding 65% and the proposed fuzzy detector showing 80%.

*4.3 Computational effort*

The speed and computational effort of a corner detector algorithm must meet demands for real-time tasks, regarding speed and required processing time. The runtime of an algorithm can be a

*11*

reference to its overall computational effort. In order to compare the three algorithms, fifty pairs of images are considered in order to register the algorithm's runtime for testing images holding 320×240 pixels. The average runtime for the Harris method, the fuzzy Banerjee & Kundu algorithm and the proposed corner detectors is 1.8686s, 6.2125s and 0.878s respectively, as all are tested under the MatLab© R2008b environment.

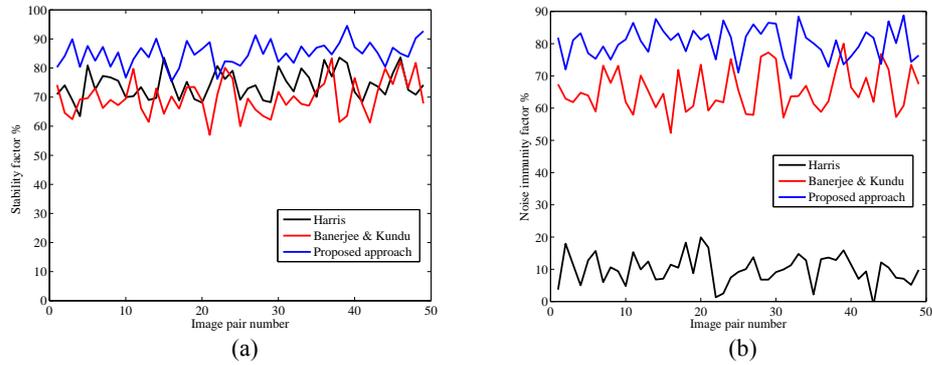

**Fig. 7.** Performance comparison among corner detectors. (a) Stability factor and (b) noisy immunity factor.

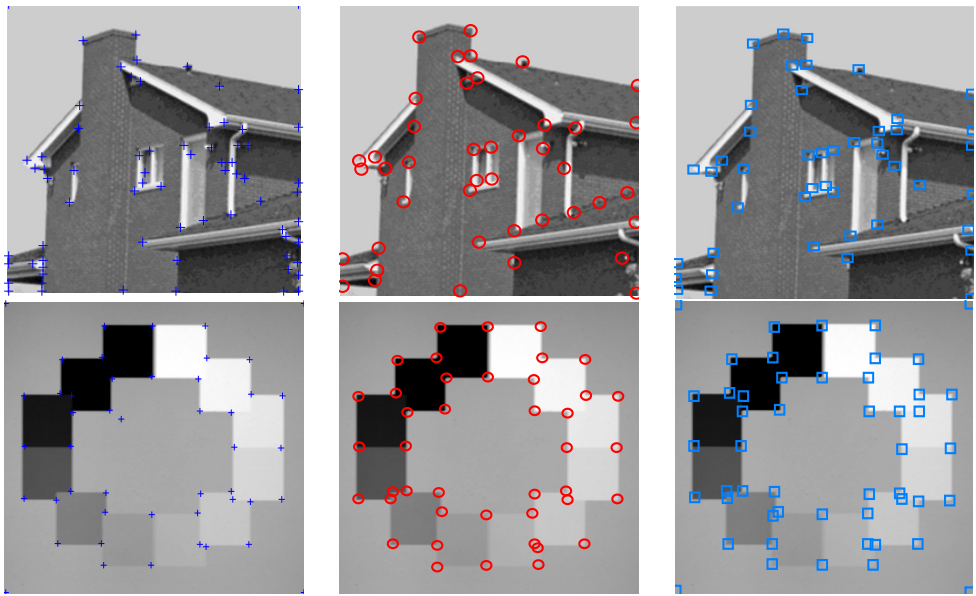

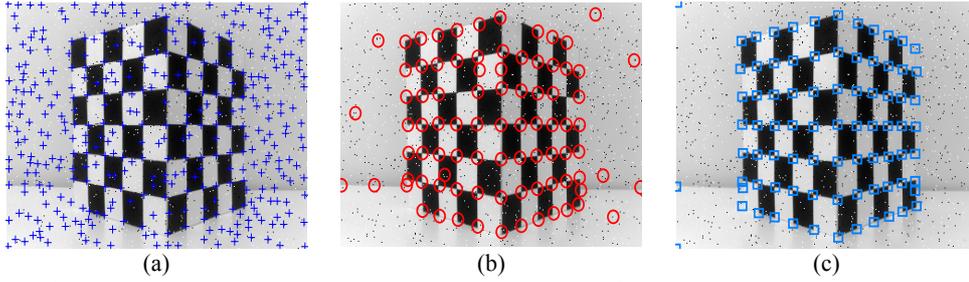

(a) (b) (c)

**Fig. 8.** Benchmark images: (a) Harris algorithm results, (b) fuzzy Banerjee & Kundu method results and (c) the proposed detector results.

*4.4 Comparison results*

Table 2 shows a final comparison between all the methods. The proposed fuzzy detector can be considered as equally stable as the Harris method. It also shows stronger noise immunity being slightly superior to the fuzzy detector proposed by Banerjee & Kundu. The proposed corner detector can also be regarded as the algorithm showing the best computational performance.

| Corner Detector | Stability± σ(%) | Noise± σ (%) | Time± σ (s) |
|---|---|---|---|
| Harris | 75± 5.5 | 9± 4.4 | 1.8686± 0.3 |
| Fuzzy Banerjee & Kundu | 70± 7.8 | 65± 7.1 | 6.2125± 0.21 |
| The proposed detector | 83± 4.1 | 80± 4.6 | 0.878± 0.11 |

**Table 2.** Performance comparison among the three corner detectors considered by the study.

Figure 8 shows the performance of the detector algorithms considered in the study while analyzing a number of benchmark images.

## 4. CONCLUSIONS

This paper has presented a corner detection algorithm which models the structure of a potential corner in an images based on a fuzzy rule set. The method is able to tolerate implicit imprecision and impulsive noise. Experimental evidence suggests that the fuzzy proposed algorithm produces better results than other common methods such as the Harris detector [10] and the fuzzy approach proposed by Banerjee & Kundu [25]. The proposed algorithm is able to successfully identify corners on images holding different uncertainty conditions. However it is also sensitive to blurring in particular when a steaming up effect is produced by considering neighborhood window wider than the one previously considered for building the fuzzy model of corners (templates). Such fact shall not be considered as inconvenient because the fuzzy-based algorithm is still capable of identifying corners over similar blurring levels than those of conventional algorithms.The proposed detector is stable and has shown robustness to impulsive noise which in turn represents its major advantage over the Harris method considering that impulsive noise is commonly found in real-time images. Although the algorithm exhibits a tolerance to imprecision that matches the performance of the Banerjee & Kundu fuzzy method, the presented approach requires a lighter computational cost for analyzing benchmark images.

# REFERENCES

1. Lowe, D.G., Perceptual Organization and Visual Recognition, Kluwer Academic Publishers, USA, 1985.
2. Loupias, E. and Sebe, E., Wavelet-based salient points: applications to image retrieval using color and texture features, in Advances in visual Information Systems, in: Proceedings of the 4th Intenational Conference, VISUAL 2000, (2000), pp. 223–232.
3. Fischler, M. and Wolf, H.C., "Locating perceptually salient points on planar curves", IEEE Trans. Pattern Anal. Mach. Intell. 16 (2) (1994) 113–129.
4. Freeman, H. and Davis, L.S., "A corner-finding algorithm for chain-coded curves", IEEE Trans. Comput. C-26 (1977) 297–303.
5. Kitchen, L. and Rosenfeld, A., "Gray-level corner detection", Pattern Recogn. Lett. 1 (1982) 95–102.
6. Zheng, Z., Wang, H. and Teoh, E., "Analysis of gray level corner detection", Pattern Recogn. Lett. 20 (2) (1999) 149–162.
7. Rattarangsi, A. and Chin, R.T., "Scale-based detection of corners of planar curves", IEEE Trans. Pattern Anal. Mach. Intell. 14 (4) (1992) 430–449.
8. Teh, C. and Chin, R.T., "On the detection of dominant points on digital curves", IEEE Trans. Pattern Anal. Mach. Intell. 11 (8) (1989) 859–872.
9. A. Rosenfeld, E. Johnston, "Angle detection on digital curves", IEEE Transaction on Computers C-22 (1973) 858–875.
10. Harris, C. and Stephens, M., A combined corner and edge detector, in: Proceedings of the 4th Alvey Vision Conference, 1988, pp. 147–151.
11. Smith, S. and Brady, M., "A new approach to low level image processing", Int. J. Comput. Vision 23 (1) (1997) 45–78.
12. Zou, L., Chen, J., Zhang, J., Dou L.: The Comparison of Two Typical Corner Detection Algorithms, Second International Symposium on Intelligent Information Technology Application ISBN 978-0-7695-3497 (2008).
13. Zadeh, L.A., Fuzzy sets, Information and Control 8 (1965) 338–353.
14. Pal, S.K., Ghosh, A. and Kundu, M.K., Soft Computing for Image Processing, Physica-Verlag, 2000, pp. 44–78 (Chapter 1).
15. Yua, D., Hu, Q. and Wua, C., "Uncertainty measures for fuzzy relations and their applications", Appl. Soft Comput. 7 (3) (2007) 1135–1143.
16. Karmakar, G., Dooley, L.: A generic fuzzy rule based in image segmentation algorithm, Pattern Recognition Letters, 23, pp. 1215-1227, (2002).
17. Basak, J., Pal, S.: Theoretical quantification of shape distortion in fuzzy Hough transform, Fuzzy Sets and Systems, 154, pp. 227-250, (2005).
18. Jacquey, F., Comby, F., Strauss, O.: Fuzzy Edge detection for omnidirectional images, Fuzzy sets and Systems, 159, pp. 1991-2010, (2008).
19. Russo, F.: FIRE operators for image processing. Fuzzy Sets and Systems, 103, pp 265-275, (1999).
20. Tizhoosh, H.: Fast and Robust Fuzzy Edge Detection, Fuzzy Filters for image processing, Nachtegael et. al. Eds. Springer, Berlin, 2003.
21. Liang, L., Looney, C.: Competitive Edge detection, Applied Soft Computing, 3, pp. 123-137, (2003).

22. Kim, D., Lee, W., Kweon, I.: Automatic edge detection using 3x3 ideal binary pixel patterns and fuzzy-based edge thresholding, Pattern Recognition Letters, 25, pp. 101-106, (2004).
23. Russo, F.: Impulse noise cancellation in image data using a two-output nonlinear filter, Measurement, 36(3-4), pp. 205-213, (2004).
24. Yüksel, M.; Edge detection in noisy images by fuzzy processing, International Journal of Electronics and Communications, 61, pp. 82-89, (2007).
25. Banerjee M. and Kundu, M. K., "Handling of impreciseness in gray level corner detection using fuzzy set theoretic approach", Applied Soft Computing, 8(4), pp. 1680-1691, (2008).
26. Várkonyi-Kóczy, A.: Fuzzy logic supported corner detection. Journal of Intelligent & Fuzzy Systems, 19, pp 41-50, (2008).
27. Moravec, H., Towards automatic visual obstacle avoidance, in: Proceedings of the 5th International Joint Conference on Artificial Intelligence, 1997, p. 584.
28. Cordelia Schmid, Rrger Mohr, Christian Bauckhage (2000). "Evaluation of interest point detectors", International Journal of Computer Vision, Vol. 37, No. 2, pp.151-172.
29. F. Mokhtarian, F. Mohanna (2006). "Performance evaluation of corner detectors using consistency and accuracy measures". Computer Vision and Image Understanding, Vol. 102, No. 1, pp. 81-94.

## ABOUT THE AUTHOR(S)

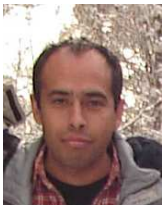

**Erik Cuevas** received the B.S. degree with distinction in Electronics and Communications Engineering from the University of Guadalajara, Mexico in 1995, the M.Sc. degree in Industrial Electronics from ITESO, Mexico in 2000, and the Ph.D. degree from Freie Universität Berlin, Germany in 2005. From 2001 he was awarded a scholarship from the German Service for Academic Interchange (DAAD) as full-time researcher. Since 2006 he has been with University of Guadalajara, where he is currently a full-time Professor in the Department of Computer Science. From 2008, he is a member of the Mexican National Research System (SNI). His research interest includes computer vision and artificial intelligence.

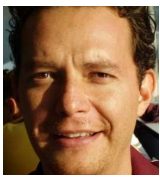

**Daniel Zaldivar** received the B.S. degree with distinction in Electronics and Communications Engineering from the University of Guadalajara, Mexico in 1995, the M.Sc. degree in Industrial Electronics from ITESO, Mexico in 2000, and the Ph.D. degree from Freie Universität Berlin, Germany in 2005. From 2001 he was awarded a scholarship from the German Service for Academic Interchange (DAAD) as full-time researcher. Since 2006 he has been with University of Guadalajara, where he is currently a Professor in the Department of Computer Science. From 2008, he is a member of the Mexican National Research System (SNI). His current research interest includes biped robots design, humanoid walking control, and artificial vision.

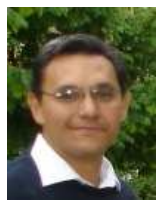

**Marco Pérez-Cisneros** received the B.S. degree with distinction in Electronics and Communications Engineering from the University of Guadalajara, Mexico in 1995, the M.Sc. degree in Industrial Electronics from ITESO University, Mexico in 2000, and the Ph.D. degree from the Control Systems Centre, UMIST, Manchester, UK in 2004. Since 2005 he has been with University of Guadalajara, where he is currently a Professor and Head of Department of Computer Science. Since 2007 he has spent yearly spells at the University of Manchester as an invited honorary Professor. He is a member of the Mexican

National Research System (SNI) from 2007. His current research interest includes robotics and computer vision in particular visual servoing applications on humanoid walking control.

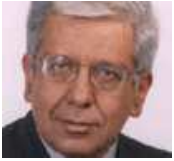

**E. N. Sanchez** obtained the BSEE from Universidad Industrial de Santander (UIS), Bucaramanga, Colombia in 1971, the MSEE from CINVESTAV-IPN (Advanced Studies and Research Center of the National Polytechnic Institute), Mexico City, Mexico, in 1974 and the Docteur Ingenieur degree in Automatic Control from Institut Nationale Polytechnique de Grenoble, France in 1980. He was granted an USA National Research Council Award as a research associate at NASA Langley Research Center, Hampton, Virginia, USA (January 1985 to March 1987). His research interest center in Neural Networks and Fuzzy Logic as applied to Automatic Control systems. He has been advisors of 6 Ph. D. thesis and 33 M. Sc Thesis. Since January 1997, he is professor of CINVESTAV_IPN, Guadalajara Campus, Mexico.

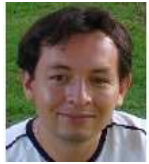

**Marte Ramírez-Ortegón** is a Ph.D. student at the Department of Mathematics and Computer Science of the Freie Universität Berlin. In 2002 he received a B.S. degree (Computer Science) from the University of Guanajuato (UG) / the Centre for Mathematical Research (CIMAT), Guanajuato, Mexico (1998–2002). His current research fields include pattern recognition, edge detection, binarization, multithresholding and text recognition